\documentclass[10pt,twocolumn,letterpaper]{article}

\usepackage{iccv}
\usepackage{times}
\usepackage{epsfig}
\usepackage{graphicx}
\usepackage{amsmath}
\usepackage{amssymb}
\usepackage{graphicx}
\usepackage{color}
\definecolor{grey}{rgb}{0.5, 0.5, 0.5}
\usepackage{multirow}
\usepackage{authblk}
\usepackage{subcaption}

\usepackage[breaklinks=true,bookmarks=false]{hyperref}

\iccvfinalcopy 

\def\iccvPaperID{****} 

\begin{document}

\title{Visual Madlibs: Fill in the blank Image Generation and Question Answering}

\title{Visual Madlibs: Fill in the blank Image Generation and Question Answering}

\author{Licheng Yu}
\author{Eunbyung Park}
\author{Alexander C. Berg}
\author{Tamara L. Berg}
\affil{Department of Computer Science, University of North Carolina, Chapel Hill
}
\affil{\tt\small \{licheng, eunbyung, acberg, tlberg\}@cs.unc.edu}

\maketitle

\begin{abstract}
In this paper, we introduce a new dataset consisting of 360,001 focused natural
language descriptions for 10,738 images.  This dataset, the Visual Madlibs
dataset, is collected using automatically produced fill-in-the-blank templates
designed to gather targeted descriptions about: people and objects, their
appearances, activities, and interactions, as well as inferences about the
general scene or its broader context. We provide several analyses of the Visual
Madlibs dataset and demonstrate its applicability to two new description
generation tasks: focused description generation, and multiple-choice
question-answering for images.  Experiments using joint-embedding and deep
learning methods show promising results on these tasks.
\end{abstract}

\begin{figure}[t!]
\centering
\includegraphics[width=0.5\textwidth]{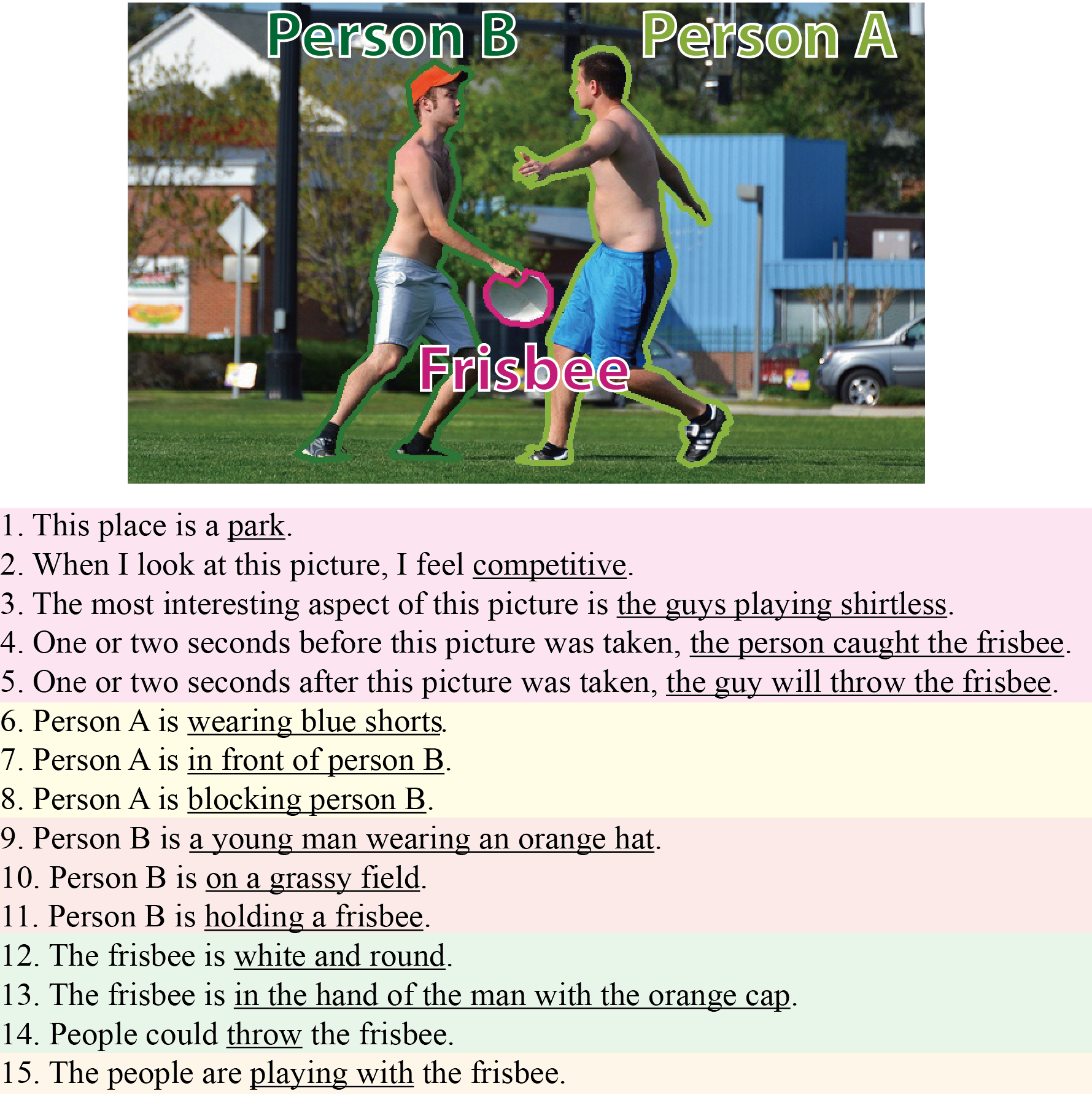}
\caption{An example from the Visual Madlibs Dataset. This dataset collects targeted descriptions for people and objects, denoting their appearances, affordances, activities, and interactions. It also provides descriptions of broader emotional, spatial and temporal context for an image.}\label{fig:frisbee}
\end{figure}

\begin{figure*}[t!]
\includegraphics[width=\textwidth]{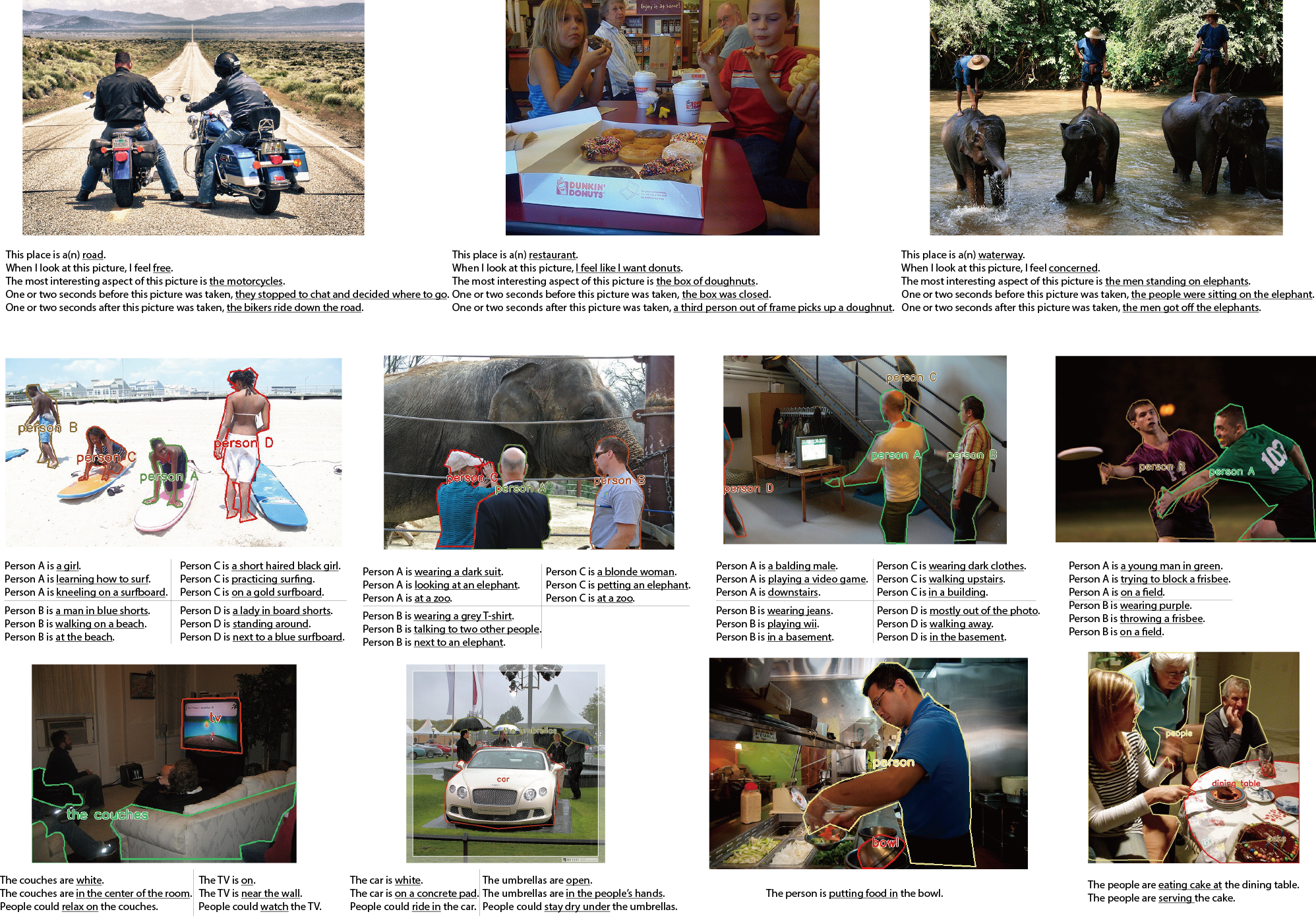}
\caption{Madlibs description. The first row corresponds to question types 1-5, the second row corresponds to question types 9-11, and the third row is to question types 6-8 and question type 12. All question types are listed in Table~\ref{table:question}.}\label{fig:more_figure}
\end{figure*}

\vspace{-.2cm}
\section{Introduction}
\vspace{-.1cm}
Much of everyday language and discourse concerns the visual world around us,
making understanding the relationship between the physical world and language
describing that world an important challenge problem for AI. Understanding this
complex and subtle relationship will have broad applicability toward inferring
human-like understanding for images, producing natural human robot interactions, and
for tasks like natural language grounding in NLP.  In computer vision, along
with improvements in deep learning based visual recognition, there has been an
explosion of recent interest in methods to automatically generate natural
language descriptions for images~\cite{chen2014learning, fang2014captions,
karpathy2014deep, vinyals2014show, kiros2014unifying,
DBLP:journals/corr/LebretPC15} or videos~\cite{venugopalan2014translating,
donahue2014long}.  However, most of these methods and existing datasets have
focused on only one type of description, a generic description for the entire
image.

In this paper, we collect a new dataset of focused, targeted, descriptions, the
{\em Visual Madlibs dataset}, as illustrated in Figure~\ref{fig:frisbee}.  To collect
this dataset, we introduce automatically produced fill-in-the-blank templates
designed to collect a range of different descriptions for visual content in an
image.  For example, a user might be presented with an image and a
fill-in-the-blank template such as ``The frisbee is [blank]'' and asked to fill
in the [blank] with a description of the appearance of frisbee. Alternatively,
they could be asked to fill in the [blank] with a description of what the
person is doing with the frisbee.  Fill-in-the-blank questions can be targeted
to collect descriptions about people and objects, their appearances,
activities, and interactions, as well as descriptions of the general scene or
the broader emotional, spatial, or temporal context of an image.  Using these
templates, we collect a large collection of 360,001 targeted descriptions for
10,738 images.
Fig.~\ref{fig:more_figure} shows some Madlibs description samples.

With this new dataset, we can develop methods to generate more focused
descriptions.  Instead of asking an algorithm to ``describe the image'' we can
now ask for more focused descriptions such as ``describe the person'',
``describe what the person is doing,`` or ``describe the relationship between
the person and the frisbee.'' We can also ask questions about aspects of an
image that are somewhat beyond the scope of the directly depicted content.  For
example, ``describe what might have happened just before this picture was
taken.'' or ``describe how this image makes you feel.'' These types of
descriptions reach toward high-level goals of producing human-like visual
interpretations for images.

In addition to focused description generation, we also introduce a
multiple-choice question-answering task for images.  In this task, the computer
is provided with an image and a partial description such as ``The person is
[blank]''.  A set of possible answers is also provided, one answer that was written
about the image in question, and several additional answers written about other images.  The
computer is evaluated on how well it can select the correct choice.  In this way,
we can evaluate performance of description generation on a concrete task,
making evaluation more straightforward.  Varying the difficulty of the negative
answers---adjusting how similar they are to the correct answer---provides a
nuanced measurement of performance.

For both the generation and question-answering tasks, we study and evaluate a
recent state of the art approach for image description
generation~\cite{vinyals2014show}, as well as a simple joint-embedding method
learned on deep representations. The evaluation also includes extensive
analysis of the Visual Madlibs dataset and comparisons to the existing MS COCO
dataset of natural language descriptions for images.

\smallskip
\noindent In summary, our contributions are:\\
1) A new description collection strategy, {\em Visual Madlibs}, for constructing fill-in-the-blank templates to collect targeted natural language descriptions.\\
2) A new Visual Madlibs Dataset consisting of 360,001 targeted descriptions, spanning 12 different types of templates, for 10,738 images, as well as analysis of the dataset and comparisons to existing MS COCO descriptions.\\
3) Evaluation of a generation method and a simple joint embedding method for targeted description generation. \\
4) Definition and evaluation of generation and joint-embedding methods on a new task, {\em multiple-choice fill-in-the-blank question answering for images}.

The rest of our paper is organized as follows. First, we review related work
(Sec~\ref{sec:related}). Then, we describe our strategy for automatically
generating fill-in-the-blank templates and introduce our Visual Madlibs dataset
(Sec~\ref{sec:dataset}). Next we outline the multiple-choice question answering and
targeted generation tasks (Sec~\ref{sec:tasks}) and provide several analyses of
our dataset (Sec~\ref{sec:analysis}). Finally, we provide experiments
evaluating description generation and joint-embedding methods on the proposed
tasks (Sec~\ref{sec:experiments}) and conclude (Sec~\ref{sec:conclusions}).

\vspace{-.2cm}
\section{Related work}
\label{sec:related}
\vspace{-.1cm}
\textbf{Description Generation:} 
Recently, there has been an explosion of interest in methods for producing
natural language descriptions for images or video.  Early work in this area
generally explored two complementary directions. The first type of approach
focused on detecting content elements such as objects, attributes, activities,
or spatial relationships and then composing captions for
images~\cite{kulkarni2011baby,yang2011corpus,mitchell2012midge,farhadi2010every} or
videos~\cite{krishnamoorthy:naacl-wvl13} using linguistically inspired
templates.  The second type of approach explored methods to make use of
existing text either directly associated with an image~\cite{feng2010topic,aker2010generating} or
retrieved from visually similar
images~\cite{ordonez2011im2text,kuznetsova2012collective,mason2013domain}.

With the advancement of deep learning for content estimation, there have been
many exciting recent attempts to generate image descriptions using neural
network based approaches. Some methods first detect words or phrases
using Convolutional Neural Network (CNN) features, then generate and re-rank
candidate sentences~\cite{fang2014captions, DBLP:journals/corr/LebretPC15}.
Other approaches take a more end-to-end approach to generate output
descriptions directly from images.  Kiros
\emph{et~al.}~\cite{kiros2014unifying} learn a joint image-sentence embedding
using visual CNNs and Long Short Term Memory (LSTM) networks. Similarly, several other methods
have made use of CNN features and LSTM or recurrent neural networks (RNN) for generation
with a variety of different
architectures~\cite{vinyals2014show,karpathy2014deep,chen2014learning}. 
These new methods have shown great promise for image description generation under some measures (e.g. BLEU-1) achieving near-human performance levels. 
We look at related, but more focused description generation tasks.

\textbf{Description Datasets:}
Along with the development of image captioning algorithms there have been a number of datasets collected for this task.  
One of the first datasets collected for this problem was the UIUC Pascal Sentence data set~\cite{farhadi2010every} which contains 1,000 images with 5 sentences per image written by workers on Amazon Mechanical Turk.  
As the description problem gained popularity larger and richer datasets were collected, including the Flickr8K~\cite{rashtchian2010collecting} and Flickr30K~\cite{young2014image} datasets, containing 8,000 and 30,000 images respectively. 
In an alternative approach, the SBU Captioned photo dataset~\cite{ordonez2011im2text} contains 1 million images with existing captions collected from Flickr. 
This dataset is larger, but the text tends to contain more contextual information since captions were written by the photo owners. 
Most recently, Microsoft released the MS COCO~\cite{DBLP:journals/corr/LinMBHPRDZ14} dataset.  MS COCO contains 120,000 images depicting 80 common object classes, with object segmentations and 5 turker written descriptions per image. 
These datasets have been one of the driving forces in improving methods for description generation, but are currently limited to a single description about the general content of an image.  
We make use of MS COCO data, extending the types of descriptions associated with images.

\textbf{Question-answering}
Natural language question-answering has been a long standing goal of NLP, with commercial companies like Ask-Jeeves or Google playing a significant role in developing effective methods.  Recently, embedding and deep learning methods have shown great promise for question-answering~\cite{sukhbaatar2015weakly,bordes2014question,bordes2014open}.
Lin \emph{et~al.}~\cite{lin2015don} take an interesting multi-modal approach to question-answering. A multiple-choice text-based question is first constructed from 3 sentences written about an image; 2 of the sentences are used as the question, and 1 is used as the positive answer, mixed with several negative answers from sentences written about other images. 
The authors develop ranking methods to answer these questions and show that generating abstract images for each potential answer can improve results. 
Note, here the algorithms are not provided with an image as part of the question.  Some recent work has started to look at the problem of question-answering for images.  
Malinowski \emph{et~al.}~\cite{malinowski2014multi} combine computer vision and NLP in a Bayesian framework, but restrict their method to scene based questions. 
Geman \emph{et~al.}~\cite{geman2015visual} design a visual Turing test to test image understanding using a series of binary questions about image content.  
We design more general question-answering tasks that allow us to ask a variety of different types of natural language questions about images.

\begin{table*}[th!]
\scriptsize
\begin{center}
\begin{tabular}{| l | l | l | c |}
\hline
 \multicolumn{1}{|c|}{\bfseries Type}& \multicolumn{1}{|c|}{\bfseries Instruction} & \multicolumn{1}{|c|}{\bfseries Prompt}  & \multicolumn{1}{|c|}{\bfseries \#words}\\
\hline \hline
1. image's scene & Describe the type of scene/place shown in this picture.&The place is a(n) \rule{0.5cm}{0.05em} .&\textcolor{grey}{4}+1.45\\
\hline
2. image's emotion& Describe the emotional content of this picture.  &When I look at this picture, I feel \rule{0.5cm}{0.05em} .&\textcolor{grey}{8}+1.14  \\
\hline
3. image's interesting& Describe the most interesting or unusual aspect of this picture.&The most interesting aspect of this picture is \rule{0.5cm}{0.05em} .&\textcolor{grey}{8}+3.14  \\
\hline
4. image's past & Describe what happened immediately before this picture was taken.  & One or two seconds before this picture was taken, \rule{0.5cm}{0.05em} .&\textcolor{grey}{9}+5.45 \\
\hline
5. image's future & Describe what happened immediately after this picture was taken. & One or two seconds after this picture was taken, \rule{0.5cm}{0.05em} .&\textcolor{grey}{9}+5.04 \\
\hline
6. object's attribute& Describe the appearance of the indicated object. & The object(s) is/are \rule{0.5cm}{0.05em} .&\textcolor{grey}{3.20}+1.62\\
\hline
7. object's affordance & Describe the function of the indicated object. & People could \rule{0.5cm}{0.05em} the object(s). &\textcolor{grey}{4.20}+1.74 \\
\hline
8. object's position& Describe the position of the indicated object. & The object(s) is/are \rule{0.5cm}{0.05em} . &\textcolor{grey}{3.20}+3.35\\
\hline
9. person's attribute & Describe the appearance of the indicated person/people. &The person/people is/are \rule{0.5cm}{0.05em} . &\textcolor{grey}{3}+2.52 \\
\hline
10. person's activity & Describe the activity of the indicated person/people. & The person/people is/are \rule{0.5cm}{0.05em} .  &\textcolor{grey}{3}+2.47\\
\hline
11. person's location & Describe the location of the indicated person/people. & The person/people is/are \rule{0.5cm}{0.05em} . &\textcolor{grey}{3.20}+3.04 \\ 
\hline
12. pair's relationship & Describe the relationship between the indicated person and object. &The person/people is/are \rule{0.5cm}{0.05em} the object(s). &\textcolor{grey}{5.20}+1.65 \\ 
\hline
\end{tabular}
\end{center}
\caption{All 12 types of Madlibs instructions and prompts. 
Right-most column shows the average number of words for each description (\textcolor{grey}{\#words for prompt} + \#words for answer).}\label{table:question}
\end{table*}

\vspace{-.2cm}
\section{Designing and collecting Visual Madlibs}
\label{sec:dataset}
\vspace{-.1cm}

The goal of Visual Madlibs is to study targeted natural language descriptions
of image content that go beyond describing which objects are in the image, and
beyond generic descriptions of the whole image.  The experiments in this paper
begin with a dataset of images where the presence of some objects have already
been labeled\footnote{More generally, acquiring such labels could be included
as part of collecting Madlibs.}.  The prompts for the Madlibs-style
fill-in-the-blank questions are automatically generated based on image
content, in a manner designed to elicit more detailed descriptions of the
objects, their interactions, and the broader context of the scene shown in each
image.\\

\noindent{\bf Visual Madlibs: Image+Instruction+Prompts+Blank}\\
A single fill-in-the-blank question consists of a prompt and a blank, e.g.,
{\em Person A is [blank] the car.}  The implicit question is, ``What goes in
the blank?''  This is presented to a person along with an image and
instructions, e.g., {\em Describe the relationship between the indicated person
and object.}  The same image and prompt may be used with different instructions
to collect a variety of description types.  

\noindent{\bf Instantiating Questions}\\ 
While the general form of the questions for the Visual Madlibs were chosen by
hand, see Table~\ref{table:question}, most of the questions are instantiated
depending on a subset of the objects present in an image.  For instance, if an
image contained two people and a dog, questions about each person (question
types 9-11 in Table~\ref{table:question}), the dog (types 6-8), relationships
between the two people and the dog (type 12), could be instantiated.  
For each possible instantiation, the wording of the questions might alter slightly to maintain grammatical consistency.  
In addition to these types of questions that depend on the objects present in the
image, other questions (types 1-5) can be instantiated for an image regardless
of the objects present.  

Notice in particular the questions about the temporal context -- what might have
happened before or what might happen after the image was taken. People can
make inferences beyond the specific content depicted in an image. Sometimes
these inferences will be consistent between people (e.g., when what will happen
next is obvious), and other times these descriptions may be less consistent. We
can use the variability of returned responses to select images for which these
inferences are reliable.

Asking questions about every object and all pairs of objects quickly becomes
unwieldy as the number of objects increases.  To combat this, we choose a
subset of objects present to use in instantiating questions.  Such selection
could be driven by a number of factors.  The experiments in this paper consider
comparisons to existing, general, descriptions of images, so we instantiate
questions about the objects mentioned in those existing natural language
descriptions. Whether an object is mentioned in an image description can be
viewed as an indication of the object's importance~\cite{importance}.

\vspace{-.1cm}
\subsection{Data Collection}\label{sec:collection}
\begin{figure}[h!]
\centering
\includegraphics[width=0.32\textwidth]{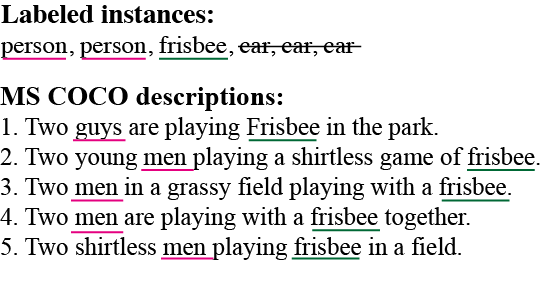}
\vspace{-.3cm}
\caption{COCO instance annotation and descriptions for the image of Fig.~\ref{fig:frisbee}. We show how we map labeled instances to the mentioned person and object in the sentence.}\label{fig:frisbee2}
\end{figure}

\vspace{-.1cm}
To collect the Visual Madlibs Dataset we use a subset of 10,738 human-centric
images from MS COCO, that make up about a quarter of the validation
data~\cite{DBLP:journals/corr/LinMBHPRDZ14}, and instantiate fill-in-the-blank
templates as described above.  The MS COCO images are annotated with a list of
objects present in the images, segmentations for the locations of those
objects, and 5 general natural language descriptions of the image.  To select
the subset of images for collecting Madlibs, we start with the 19,338 images
with a person labeled.  We then look at the five descriptions for each and
perform a dependency parse~\cite{de2006generating}, only keeping those images
where a word referring to a person (woman, man, etc. E.g., in
Fig.~\ref{fig:frisbee2}, guys, men) is the head noun for part of the parse.
This leaves 14,150 images.  We then filter out the images whose descriptions do
not include a synonym for any of the 79 non-person object categories labeled in
the MS COCO dataset.  This leaves 10,738 human-centric images with at least one
other object from the MS COCO data set mentioned in the general image
descriptions.

Before final instantiation of the fill-in-the blank templates, we need to
resolve a potential ambiguity regarding which objects are referred to in the
descriptions.  There could be several different people or different instances
of an object type labeled in an image.  It is not immediately obvious which ones are
described in the sentences.  To address this assignment problem, we estimate
the quantity of each described person/object in the sentence by parsing the
determinant (two men and a frisbee in Fig.~\ref{fig:frisbee2}), the conjunction
(a man and a woman), and the singular/plural form (dog, dogs).  We compare this
number with the number of annotated instances for each category, and consider
two possible cases: 1) there are fewer annotated instances than the sentences
describe,  2) there are more annotated instances than the sentences describe.
It is easy to address the first case, just construct templates for all of the
labeled instances.  For the second case, we sort the area of each segmented
instance, and pick the largest ones up to the parsed number for
instantiation.  Using this procedure, we obtain 26,148 labeled object
or person instances in the 10,738 images. 

Each Visual Madlib is answered by 3 workers on Amazon's Mechanical Turk.  
To date, we have collected 360,001 answers to Madlib questions.
Some example Madlibs answers are shown in Fig.~\ref{fig:more_figure}, 

\vspace{-.4cm}
\section{Tasks: Multiple-choice question answering and targeted generation}\label{sec:tasks}
\vspace{-.1cm}

We design two tasks to evaluate targeted natural language description for
images.  The first task is to automatically generate natural language descriptions of
images to fill in the blank for one of the Madlibs questions. This allows for producing
targeted descriptions such as: a description specifically focused on the appearance of an object, or
a description about the relationship between two objects.  The input to this task is an
image, instructions, and a Madlibs prompt.  As has been discussed at length in
the community working on description generation for images, it can be difficult
to evaluate free form generation.  Our second task tries to address this issue by
developing a new targeted multiple-choice question answering task for images.
Here the input is again an image, instruction, and a prompt, but instead of a
free form text answer, there are a fixed set of multiple-choice answers to fill
in the blank.  The possible multiple-choice answers are sampled from the
Madlibs responses, one that was written for the particular
image/instruction/prompt as the correct answer, and distractors chosen from
either similar images or random images depending on the level of difficulty
desired.  This ability to choose distractors to adjust the difficulty of the
question as well as the relative ease of evaluating multiple choice answers are
attractive aspects of this new task.

In our experiments we randomly select 20\% of the 10,738 images to use as our
test set for evaluating these tasks.  For the multiple-choice questions we form
two sets of answers for each, with one set designed to be more difficult than
the other.  We first establish the easy task distractor answers by randomly
choosing three descriptions (of the same question type) from other
images~\cite{lin2015don}.  The hard task is designed more delicately.  Instead
of randomly choosing from the other images, we now only look for those
containing the same objects as our question image, and then arbitrarily pick
three of their descriptions.  Sometimes, the descriptions sampled from
``similar'' images could also be good answers for our questions (later we
experiment with using Turkers to select less ambiguous multiple-choice
questions from this set).  For the targeted generation task, for question types
1-5, algorithms generate descriptions given the image, instructions, and
prompt.  For the other question types whose prompts are related to some
specific person or object, we additionally provide the algorithm with the
location of each person/object mentioned in the prompt. We also experiment with
estimating these locations using object detectors.


\vspace{-.2cm}
\section{Analyzing the Visual Madlibs Dataset}\label{sec:analysis}
\vspace{-.1cm}

We begin by conducting quantitative analyses of the responses collected in the
Visual Madlibs Dataset in Sec.~\ref{sec:numbers}. A main goal is understanding
what additional information is provided by the targeted descriptions in the
Visual Madlibs Dataset vs general image descriptions.  The MS COCO
dataset~\cite{DBLP:journals/corr/LinMBHPRDZ14} collects general image
descriptions following a similar methodology to previous efforts for collecting
general image descriptions,
e.g.~\cite{rashtchian2010collecting,young2014image}.  So, we provide further
analyses comparing the Visual Madlibs to the MS COCO descriptions collected for
the same images in Sec.~\ref{section_data}


\begin{figure*}[t!]
\centering
\includegraphics[width=0.22\textwidth]{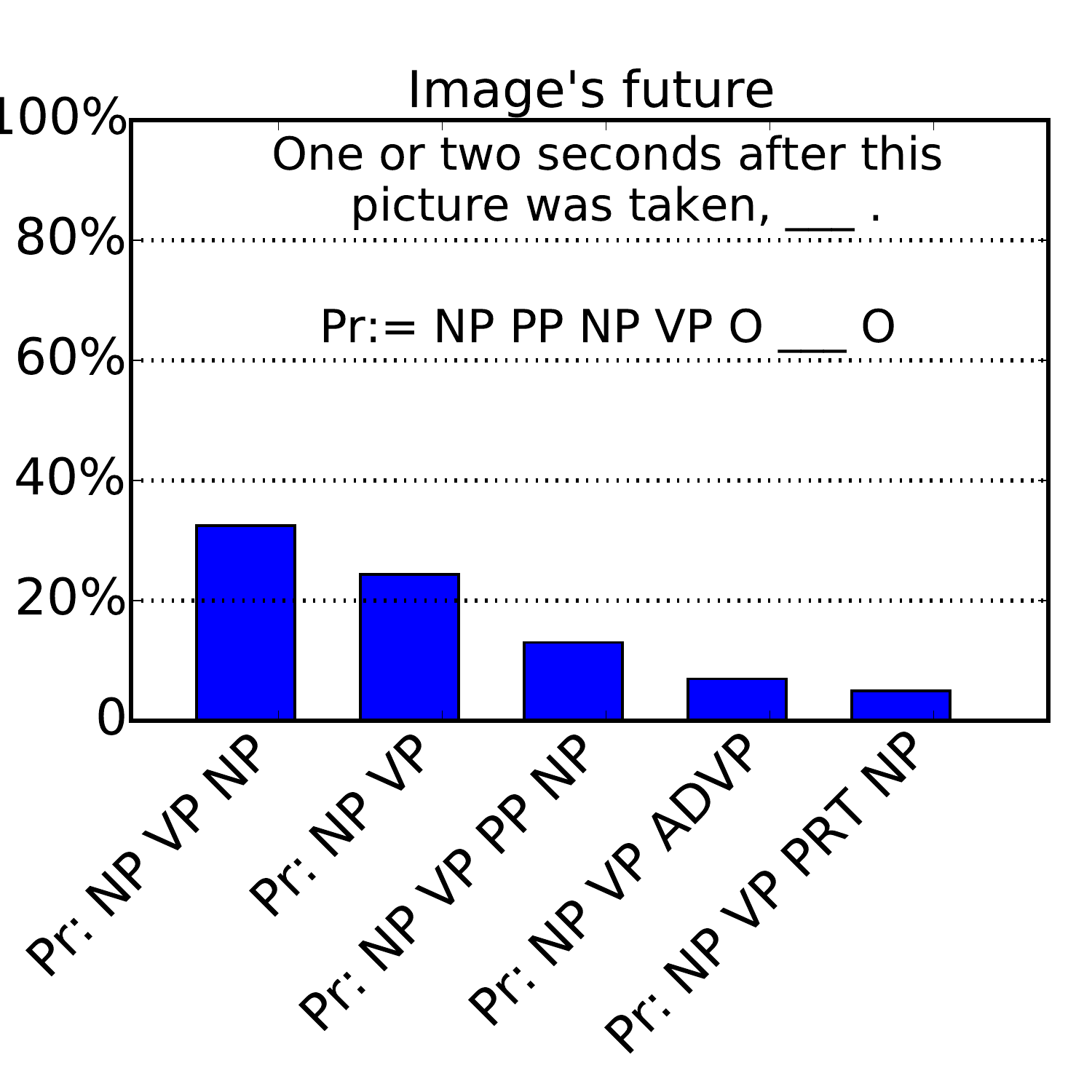}
\includegraphics[width=0.22\textwidth]{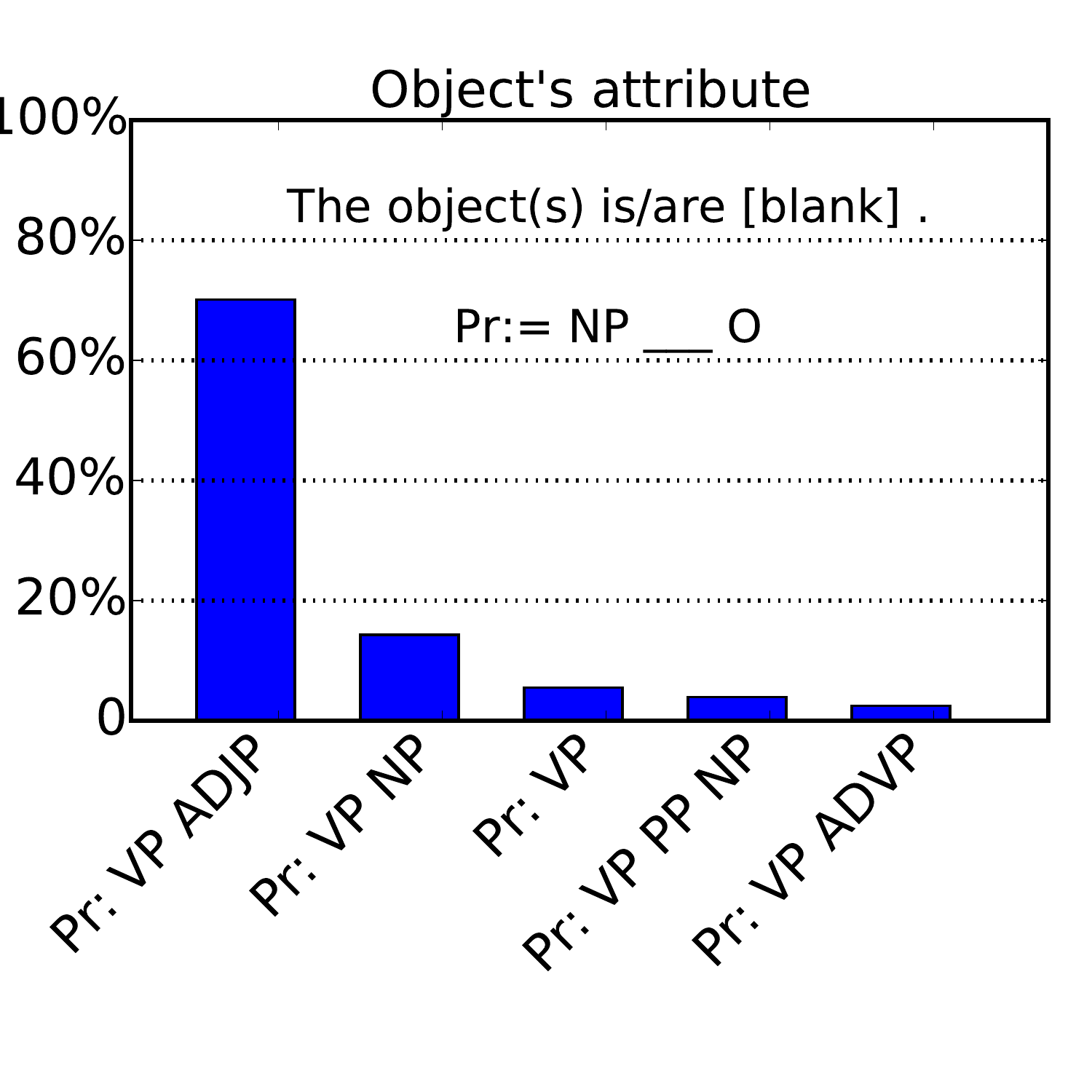}
\includegraphics[width=0.22\textwidth]{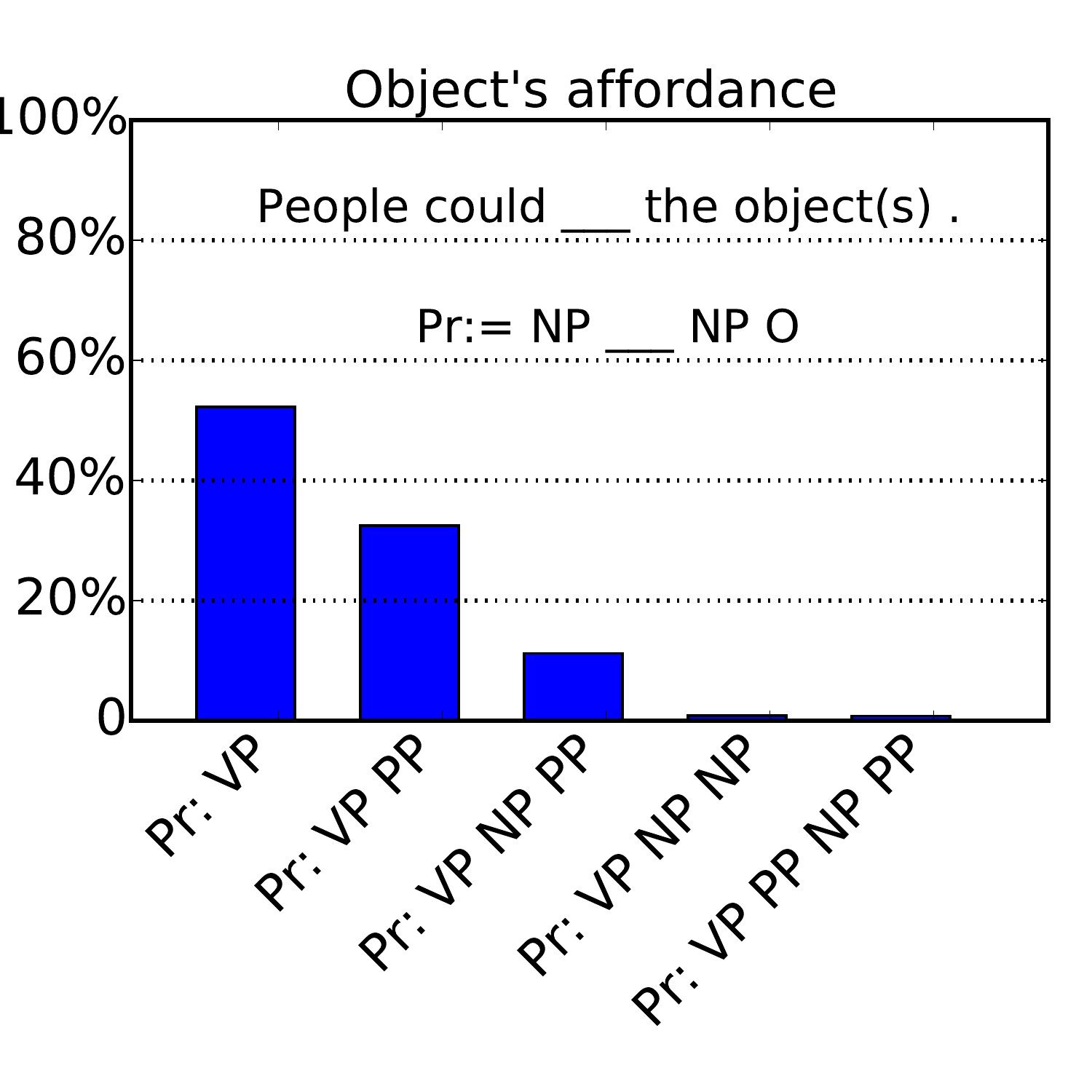}
\includegraphics[width=0.22\textwidth]{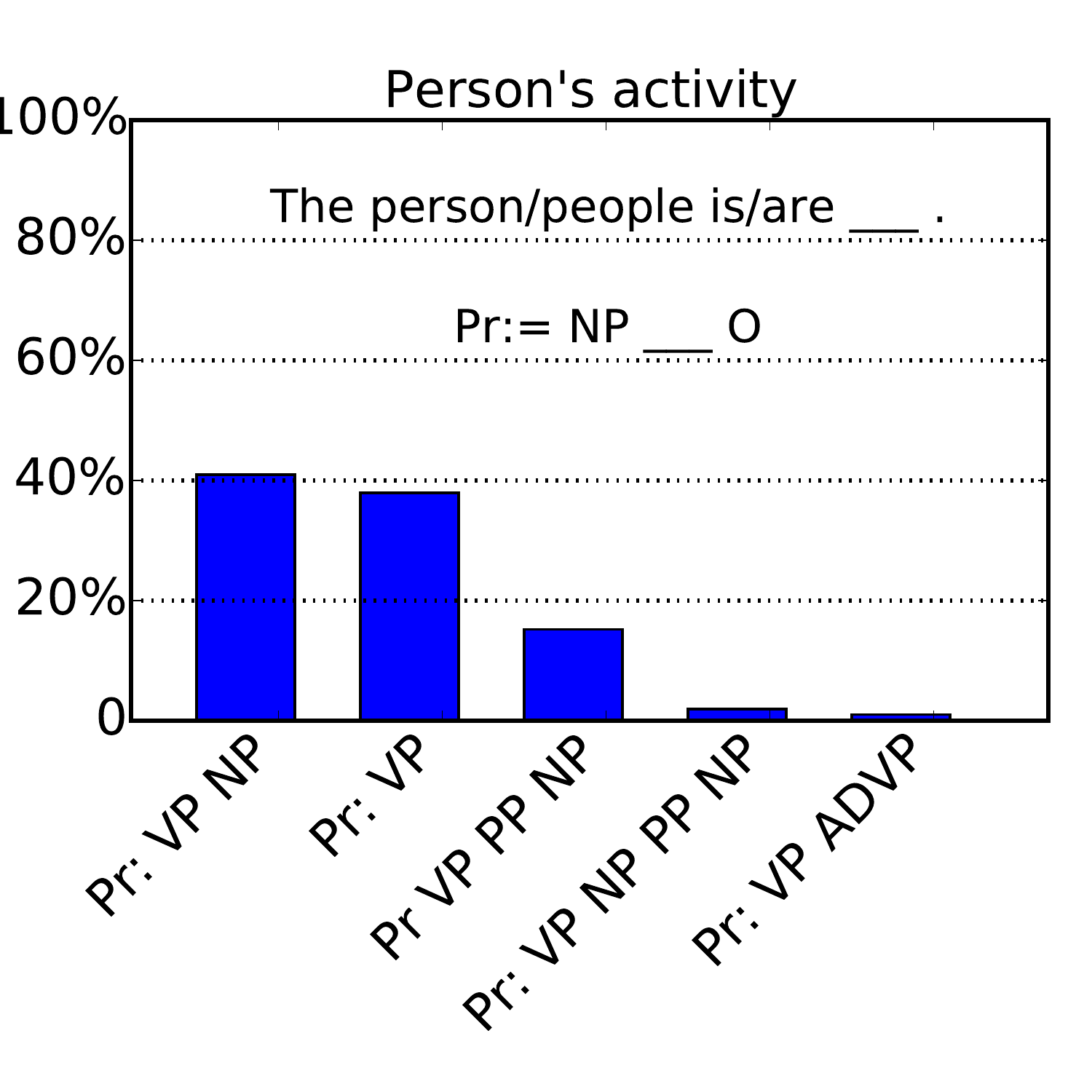}\\\vspace{-0.2cm}
\includegraphics[width=0.22\textwidth]{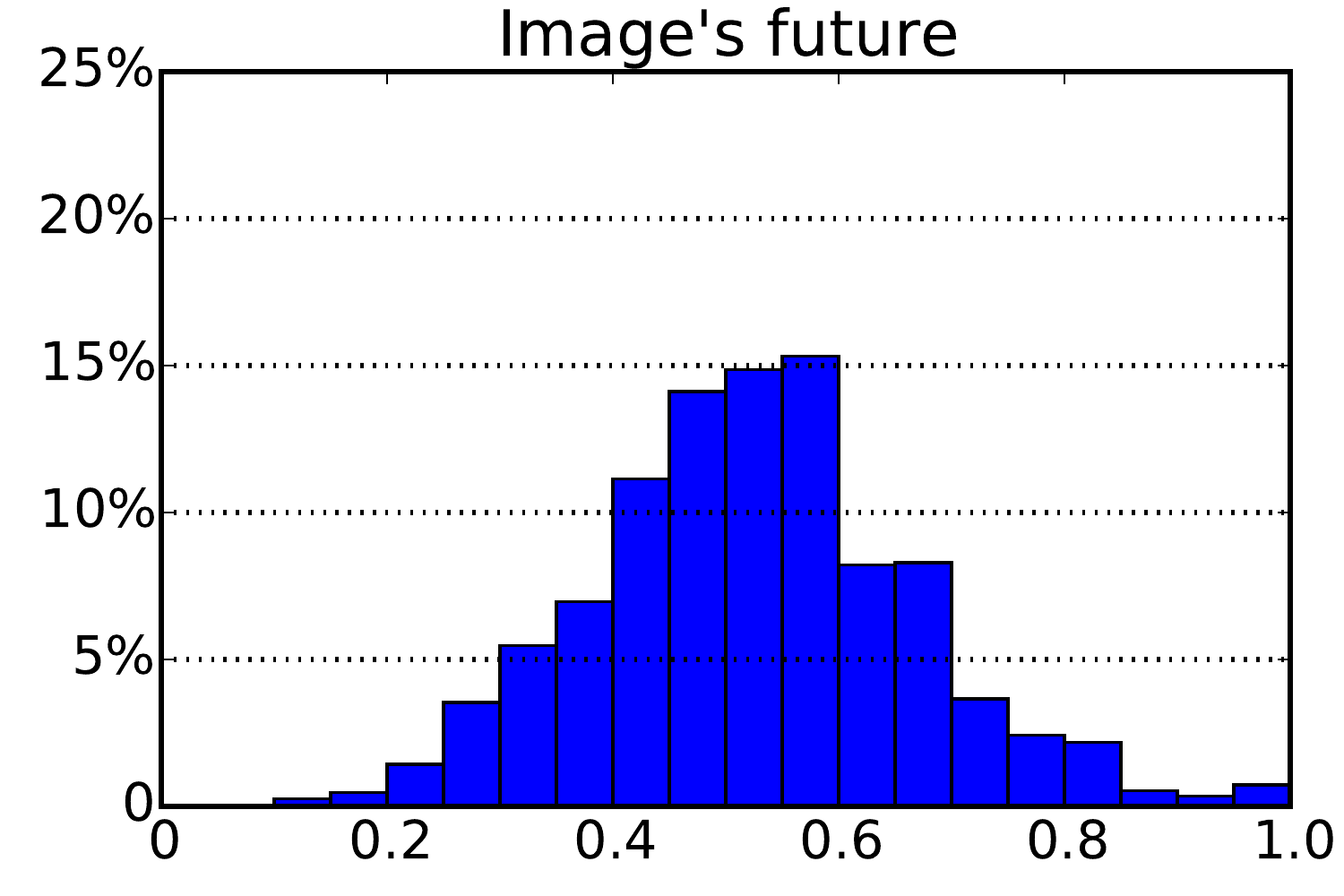}
\includegraphics[width=0.22\textwidth]{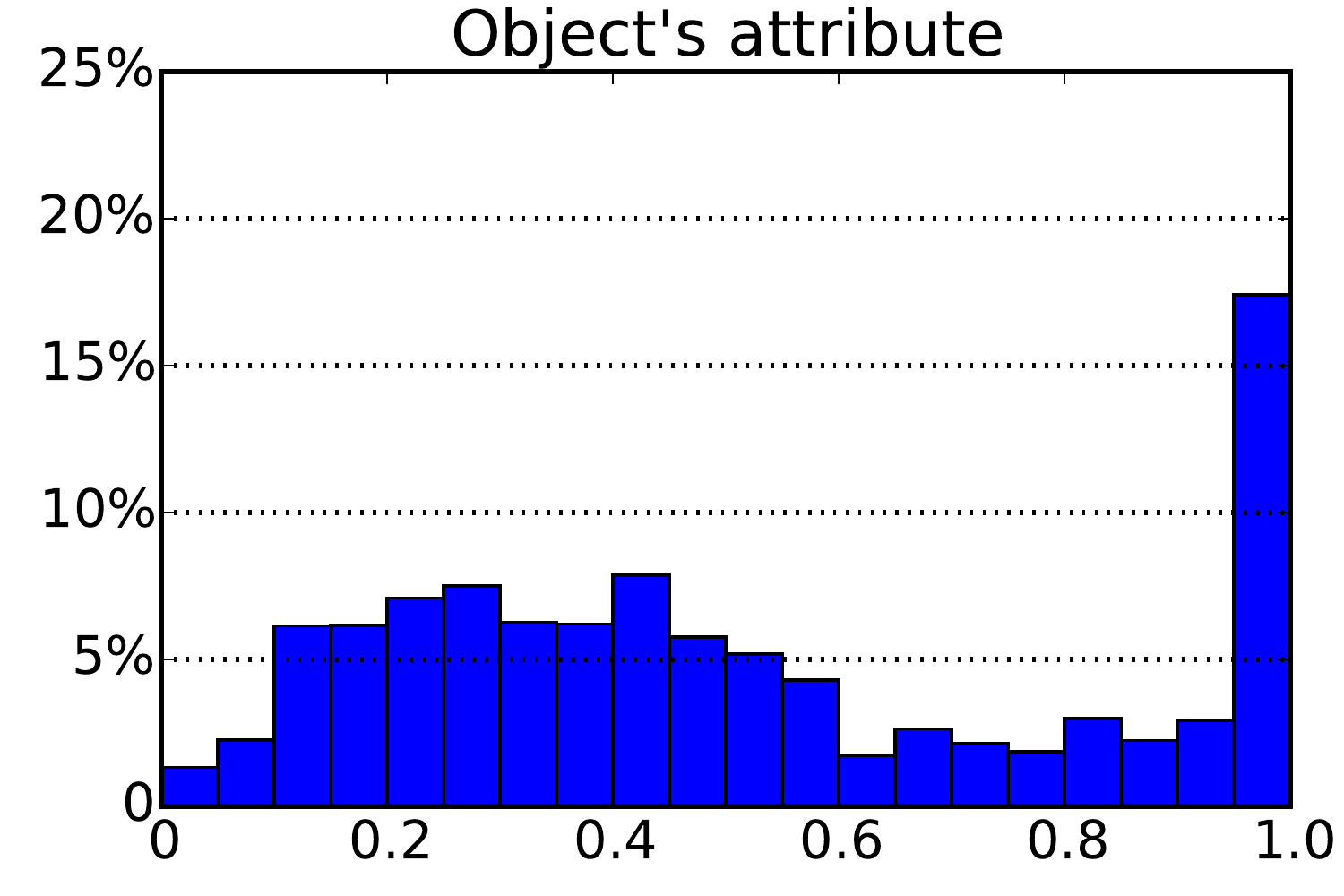}
\includegraphics[width=0.22\textwidth]{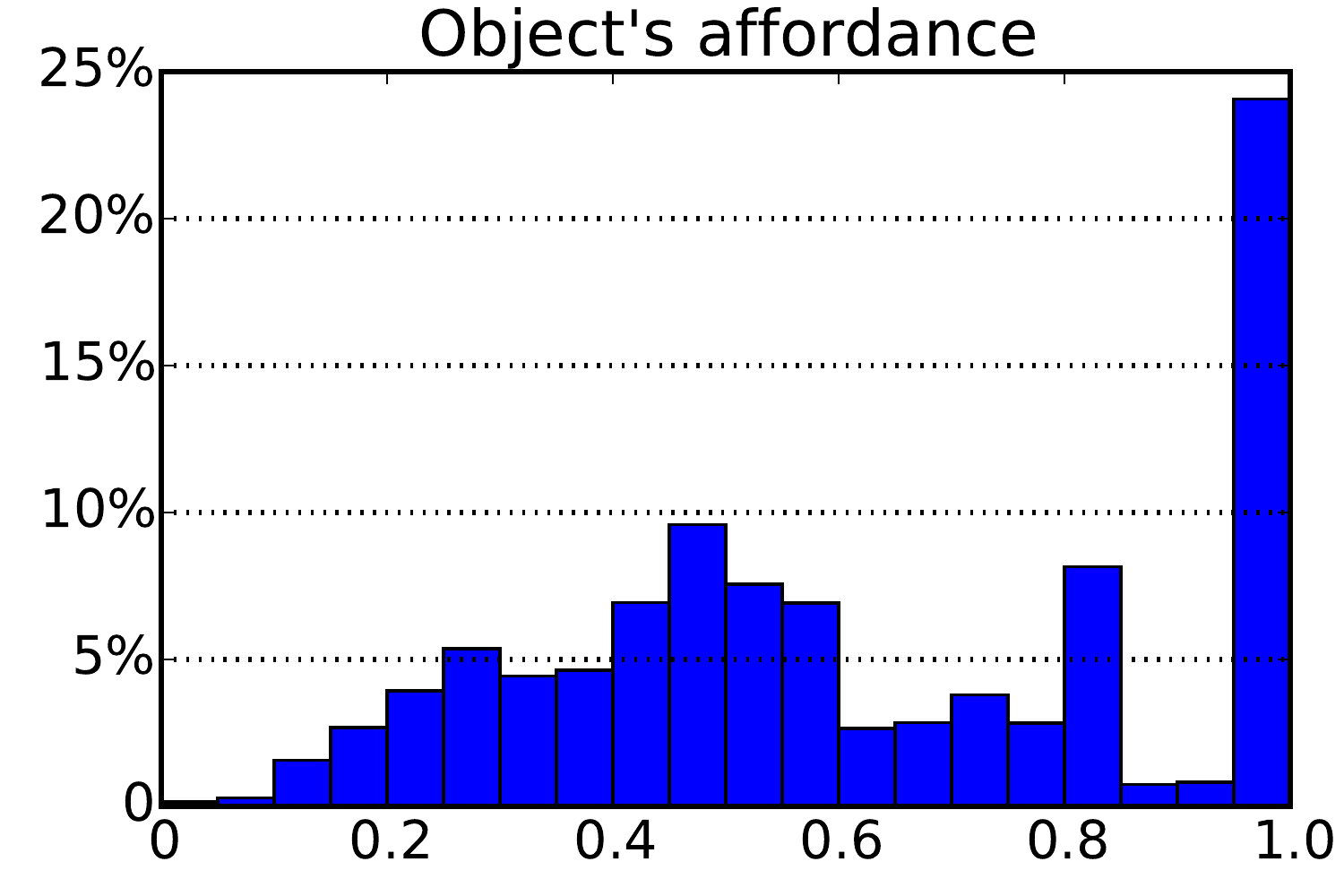}
\includegraphics[width=0.22\textwidth]{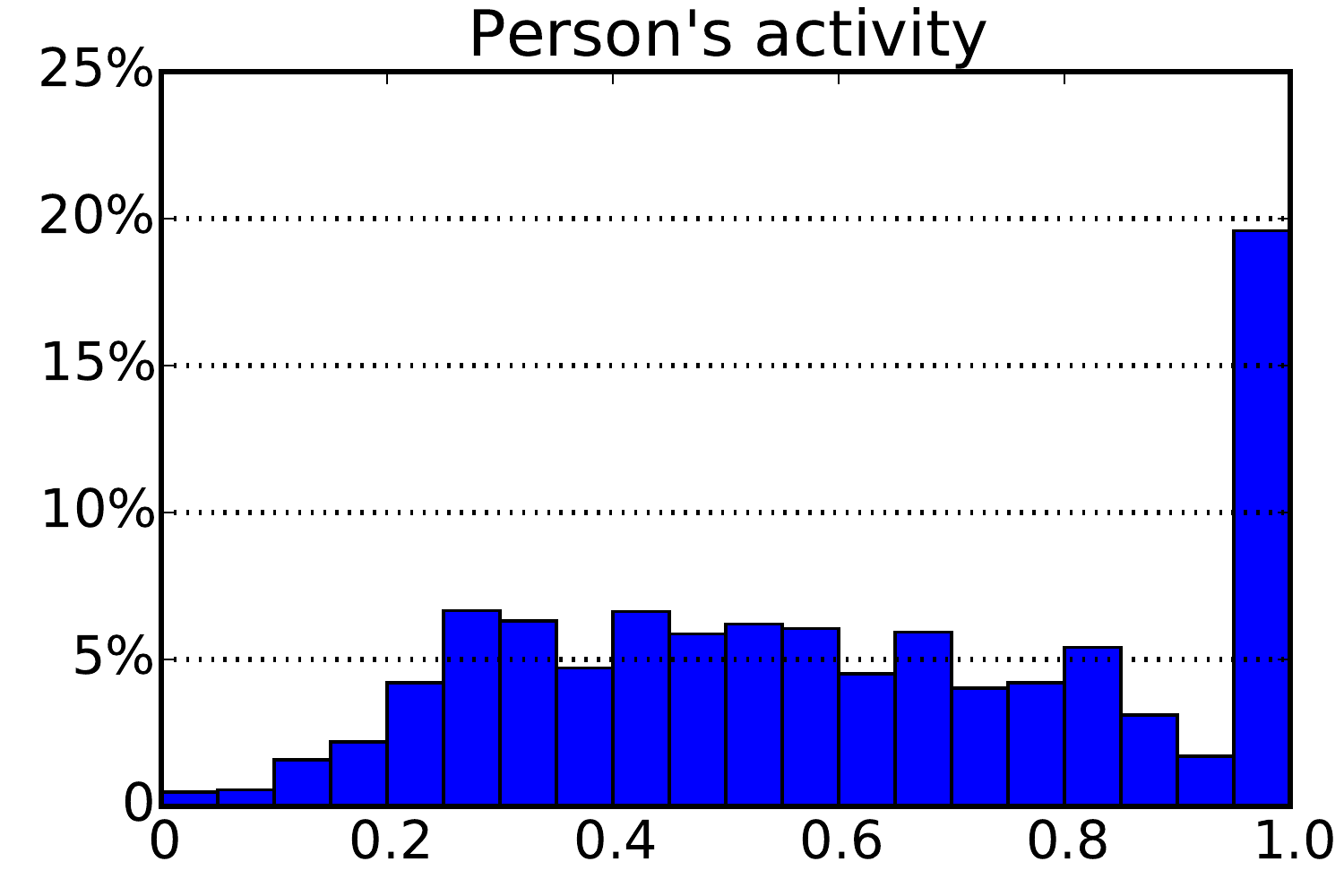}
\caption{First row shows top-5 most frequent phrase templates for image's future, object's attribute, object's affordance and person's activity. Second row shows the histograms of similarity between answers.}\label{fig:tpl_sim}
\end{figure*}

\vspace{-.2cm}
\subsection{Quantifying Visual Madlibs responses}\label{sec:numbers}
\vspace{-.1cm}

We analyze the length, structure, and consistency of the Visual Madlibs
responses.  First, the average length of each type of description is shown in
the far right column of Table~\ref{table:question}.  Note that descriptions of
people tend to be longer than descriptions of other objects in the
dataset\footnote{Also note that the length of the prompts varies slightly
depending on the object names used to instantiate the Madlib, hence the
fractional values in the mean length of the prompts shown in gray.}.

Second, we use the phrase chunking~\cite{collobert2011natural} to analyze which
phrasal structures are commonly used to fill in the blanks for different
questions.  Fig.~\ref{fig:tpl_sim}, top row, shows relative frequencies for the
top-5 most frequent templates used for several question types.  Object
attributes are usually described briefly with a simple adjectival phrase.  On
the other hand, people use more words and a wider variety of structure to
describe possible future events.  Except for 
future and past descriptions, the distribution of structures is generally concentrated
on a few likely choices for each question type. 

Third, we analyze how consistent the Mechanical Turk workers' answers are for
each type of question.  To compute a measure of similarity between a pair of
responses we use the cosine similarity between representations of each
response.  A response is represented by the mean of the
Word2Vec~\cite{mikolov2013efficient} vectors for each word in the response,
following~\cite{lin2015don,DBLP:journals/corr/LebretPC15}.  Word2Vec is a 300
dimensional embedding representation for words that encodes the distributional
context of words learned over very large word corpora.  This measure takes into
account the actual words used in a response, as opposed to the previous
analyses of parse structure.  Each Visual Madlibs question is answered by three
workers, providing 3 pairs for which similarity is computed.
Fig.~\ref{fig:tpl_sim}, bottom row, shows a histogram of all pairwise
similarities for several question types.  Generally the similarities have a
normal-like distribution with an extra peak around 1 indicating the fraction of
responses that agree almost perfectly.  Once again, descriptions of the future
and past are least likely to be (near) identical, while object attributes and
affordances are often very consistent.

\vspace{-.1cm}
\subsection{Visual Madlibs vs general descriptions}\label{section_data}
\vspace{-.1cm}


We compare the targeted descriptions in the Visual Madlibs Dataset to the
general image descriptions in MS COCO.
First, we analyze the words used in Visual Madlibs compared to MS COCO
descriptions of the same images.  For each image, we extract the unique set of
words from all descriptions of that image from both datasets, and compute the
coverage of each set with respect to the other.  We find that on average
(across images) 22.45\% of the Madlibs's words are also present in MSCOCO
descriptions, while 52.38\% of the COCO words are also present in Madlibs.

\begin{figure}[t!]
\centering
\includegraphics[width=0.45\textwidth]{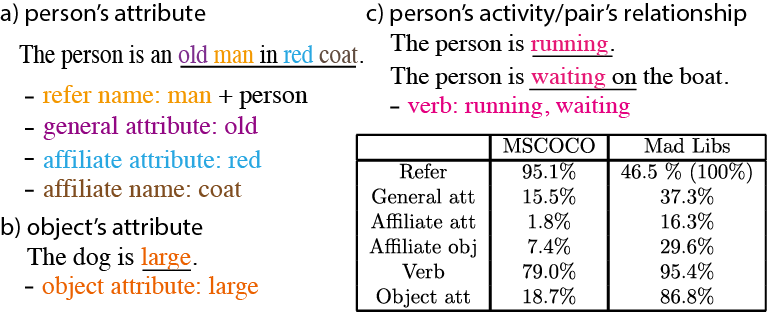}
\caption{Template used for parsing person's attributes, activity and interaction with object, and object's attribute. The percentages below compares Madlibs and MSCOCO on how frequent these templates are used for description.}\label{fig:person_att}
\end{figure}

\begin{figure}[t!]
\centering
\includegraphics[width=0.35\textwidth]{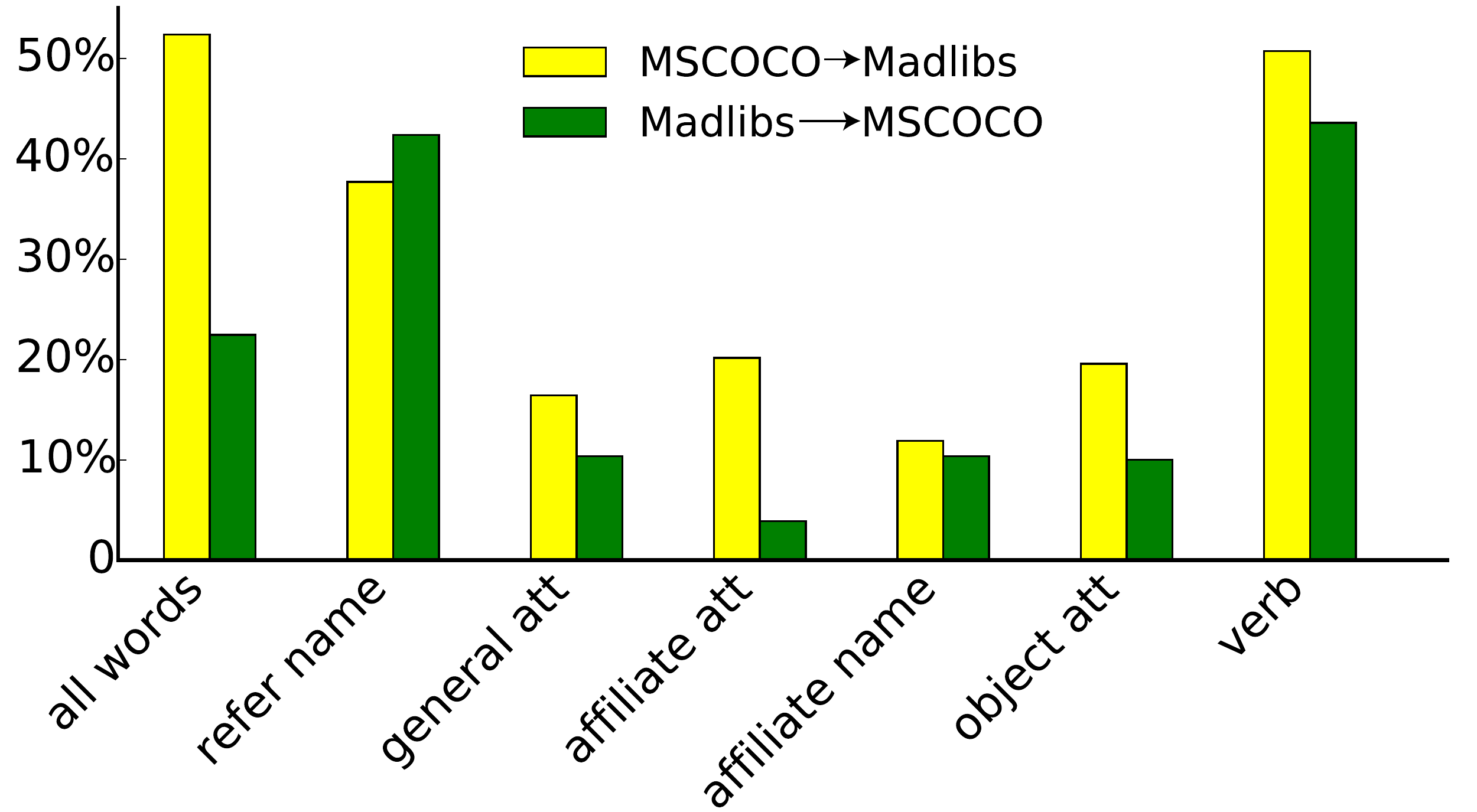}
\caption{Frequency that a word in a position in the people and object parsing template in one dataset is in the same position for the other dataset.} \label{fig:precision}
\end{figure}

Second, we compare how Madlibs and MS COCO answers describe the people and
objects in images.  We observe that the Madlibs questions types,
Table~\ref{table:question}, cover much of the information in MS COCO
descriptions~\cite{DBLP:journals/corr/LebretPC15}. As one way to see this, we
run the StanfordNLP parser\footnote{\url{http://nlp.stanford.edu/software/lex-parser.shtml}} on
both datasets.  For attributes of people, we use the parsing template shown in
Fig.~\ref{fig:person_att}(a) to analyze the structures being used.  The {\em
refer name} indicates whether the person was mentioned in the description.
Note that the Madlibs descriptions always have one reference to a person in the
prompt (The person is [blank].).  Therefore, for Madlibs, we report the
presence of additional references to the person (e.g., the person is a {\em
man}).  The {\em general
attribute} directly describes the appearance of the person or object (e.g., old
or small); the {\em affiliate object} indicates whether additional objects are
used to describe the targeted person (e.g. with a bag, coat, or glasses) and
the {\em affiliate attribute} are appearance characteristics of those secondary
objects (e.g., red coat). 
The templates for {\em object's attribute} and {\em verbs} are more straightforward as shown in Fig.~\ref{fig:person_att}(b)(c).
The table in Fig.~\ref{fig:person_att} shows the frequency of
each parse component.  Overall, more of the potential descriptive elements in
these constructions are used in response to the Madlibs prompts than in the
general descriptions found in MS COCO.  

We also break down the overlap between Visual Madlibs and MS COCO descriptions
over different parsing templates for descriptions about people and object
(Fig.~\ref{fig:precision}). Yellow bars show how often words for each parse
type in MSCOCO descriptions were also found in the same parse type in the
Visual Madlibs answers, and green bars measure the reverse direction.
Observations indicate that Madlibs provides more coverage in its descriptions
than MS COCO for all templates except for person's refer name.  One possible
reason is that the prompts already indicates ``the person'' or ``people''
explicitly, so workers need not add an additional reference to the person in their
descriptions.

\begin{figure}[t!]
\centering
\includegraphics[width=0.35\textwidth]{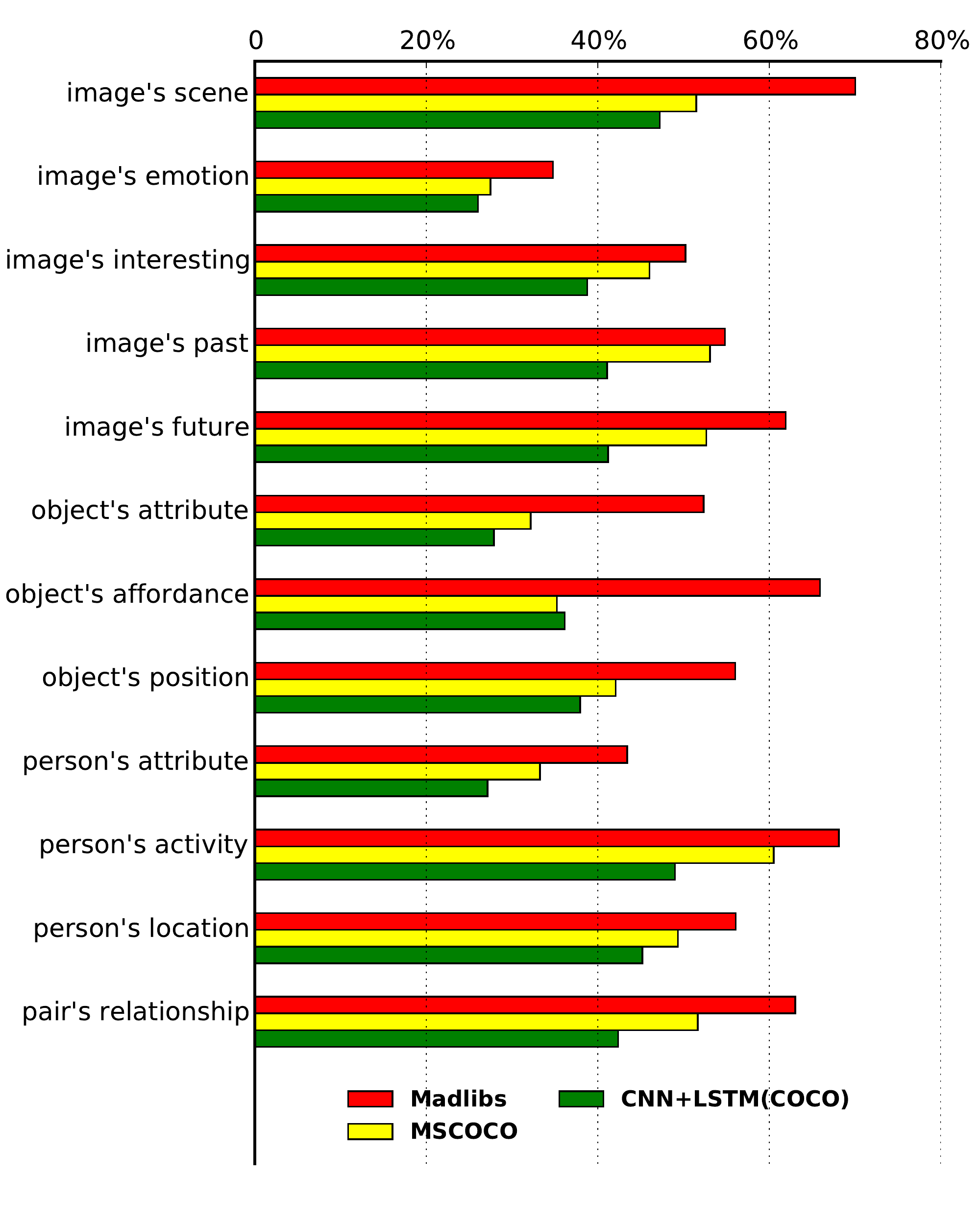}
\caption{The accuracy of Madlibs, MS COCO and CNN+LSTM~\cite{vinyals2014show}(trained on MS COCO) used as references to answer the Madlibs hard multiple-choice questions.} \label{fig:madlibs_mscoco_lstm}
\end{figure}

\noindent{\bf Extrinsic comparison of Visual Madlibs Data and general descriptions:}
Here we provide an extrinsic analysis of the information available in the
general descriptions compared to Visual Madlibs. We perform this analysis by
using either: a) the MS COCO descriptions for an image, or b) Visual Madlibs
responses from other Turkers for an image, to select answers for our
multiple-choice evaluation task. Specifically, we use one of the human provided
descriptions, either from Madlibs or from MS COCO, and select the
multiple-choice answer that is most similar to that description.  Similarity is
measured as cosine similarity between the mean Word2Vec vectors for the words a
description compared to the Word2Vec vectors of the multiple-choice answers.
In addition to comparing how well the Madlibs or MS COCO descriptions can
select the correct multiple-choice answer, we also use the descriptions
automatically produced by a recent natural language generation system
(CNN+LSTM~\cite{vinyals2014show}, implementation from~\cite{karpathy2014deep})
trained on MS COCO dataset. This allows us to make one possible measurement of
how close current automatically generated image descriptions are to our Madlibs
descriptions. Fig.~\ref{fig:madlibs_mscoco_lstm} shows
the accuracies resulting from using Madlibs, MSCOCO, or
CNN+LSTM~\cite{vinyals2014show} to select the correct multiple-choice answer.

\begin{table}
\centering
\scriptsize
\begin{tabular}{| l | c | c  c  c  c  c |}
\multicolumn{7}{c}{\bfseries Easy Task}\\ \hline
& \multicolumn{1}{c|}{\multirow{2}{*}{\#Q}} & \multicolumn{1}{c}{\multirow{2}{*}{CCA}} & \multirow{2}{*}{nCCA} & nCCA & nCCA & CNN+LSTM \\ &&&& (bbox) & (all) & (madlibs) \\ \hline
1. scene &6277& 75.7\% & 86.8\% & $-$ & \textbf{87.6}\% & 71.1\% \\ \hline
2. emotion& 5138 & 41.3\%&\textbf{49.2}\% & $-$ & 42.4\% & 34.0\% \\ \hline
3. past& 4903 & 61.8\%& 77.5\% & $-$ & \textbf{80.3}\% & 35.8\% \\ \hline
4. future& 4658&61.2\%& 78.0\% & $-$ & \textbf{80.2}\% & 40.0\% \\ \hline
5. interesting& 5095 & 66.8\% & 76.5\% & $-$ & \textbf{78.9}\% & 39.8\% \\ \hline
6. obj attr & 7194& 44.1\% & 47.5\% & \textbf{54.7}\% & 50.9\% & 45.4\% \\ \hline
7. obj aff & 7326& 59.8\%& 73.0\% & 72.2\% & \textbf{76.7}\% & $-$  \\ \hline
8. obj pos &7290& 53.0\%& 65.9\% & 58.9\% & \textbf{69.7}\% & 50.9\% \\ \hline
9. per attr& 6651& 40.4\%& 48.0\% & \textbf{53.1}\% & 44.5\% & 37.3\% \\ \hline
10. per act& 6501&70.0\%& 80.7\% & 75.6\% & \textbf{82.8}\%  & 63.7\% \\ \hline
11. per loc &6580&69.8\%& 82.7\% & 73.8\% & \textbf{82.7}\% & 59.2\% \\ \hline
12. pair rel & 7595&54.3\%& 63.0\% & 64.2\% & \textbf{67.2}\% & $-$ \\ \hline
\end{tabular}

\centering
\scriptsize
\begin{tabular}{| l | c | c  c  c  c  c |}
\multicolumn{7}{c}{\bfseries Hard Task}\\ \hline
& \multicolumn{1}{c|}{\multirow{2}{*}{\#Q}} & \multicolumn{1}{c}{\multirow{2}{*}{CCA}} & \multirow{2}{*}{nCCA} & nCCA & nCCA & CNN+LSTM \\ &&&& (bbox) & (all) & (madlibs) \\ \hline
1. scene &6277& 63.8\% &\textbf{70.1}\% & $-$ & 68.2\% & 60.5\%\\\hline
2. emotion& 5138 & 33.9\% &\textbf{37.2}\% & $-$ & 33.2\%  & 32.7\%\\\hline
3. past& 4903 & 47.9\% & 52.8\% & $-$ & \textbf{54.0}\%  & 32.0\% \\\hline
4. future& 4658& 47.5\% & \textbf{54.3}\% & $-$ & 53.3\% & 34.3\% \\\hline
5. interesting& 5095 & 51.4\%& 53.7\% & $-$ & \textbf{55.1}\%  & 33.3\% \\\hline
6. obj attr & 7194& 42.2\% & 43.6\% & \textbf{49.8}\% & 39.3\% & 40.3\%\\\hline
7. obj aff & 7326& 54.5\% & 63.5\% & \textbf{63.0}\% & 48.5\%  & $-$\\\hline
8. obj pos &7290& 49.0\% & \textbf{55.7}\% & 50.7\% & 53.4\%  & 44.9\%\\\hline
9. per attr& 6651& 33.9\% & 38.6\% & \textbf{46.1}\% & 31.6\% & 36.1\% \\\hline
10. per act& 6501& 59.7\% & 65.4\% & 65.1\% & \textbf{66.6}\%  & 53.6\%\\\hline
11. per loc &6580& 56.8\% & \textbf{63.3}\% & 57.8\% & 62.6\% & 49.3\% \\\hline
12. pair rel & 7595& 49.4\%& 54.3\% & \textbf{56.5}\% & 52.0\% & $-$ \\\hline
\end{tabular}

\centering
\scriptsize
\begin{tabular}{| l | c | c  c  c  c  c | }
\multicolumn{7}{c}{\textbf{Filtered questions} from \textbf{Hard}}\\ \hline
& \multicolumn{1}{c|}{\multirow{2}{*}{\#Q}} & \multicolumn{1}{c}{\multirow{2}{*}{CCA}} & \multirow{2}{*}{nCCA} & nCCA & nCCA & CNN+LSTM \\ &&&& (bbox) & (all) & (madlibs) \\ \hline
1. scene & 4938 & 70.4\% & \textbf{77.6}\% & $-$ & 76.3\% & 66.3\% \\\hline
2. emotion & 1936 & 43.7\% & \textbf{49.4}\% & $-$ & 44.2\% & 34.5\% \\\hline
3. future & 3628 & 52.0\% & \textbf{60.2}\%& $-$ & 59.4\% & 33.4\% \\\hline
4. past & 3811 & 51.8\% & 58.0\% & $-$ & \textbf{60.1}\% & 31.1\% \\\hline
5. interesting & 4061 & 56.5\% & 60.1\% & $-$ & \textbf{61.7}\% & 35.5\% \\\hline
6. obj attr & 5313 & 45.3\% & 47.1\% & \textbf{54.5}\% & 43.0\% & 43.4\% \\\hline
7. obj aff & 3829 & 62.7\% & \textbf{72.3}\% & 72.0\% & 59.2\% & $-$ \\\hline
8. obj pos & 5240 & 53.8\% & \textbf{61.2}\% & 55.5\% & 58.6\% & 47.6\% \\\hline
9. per attr & 4887 & 36.5\% & 42.4\% & \textbf{52.2}\% & 34.4\% & 37.1\% \\\hline
10. per act & 5707 & 62.1\% & 68.6\% & 68.1\% & \textbf{69.9}\% & 54.9\% \\\hline
11. per loc & 4992 & 63.2\% & 70.2\% & 63.0\% & \textbf{70.3}\% & 51.5\% \\\hline
12. pair rel & 5976 & 52.2\% & 57.6\% & \textbf{60.0}\% & 56.5\% & $-$ \\\hline
\end{tabular}
\caption{Accuracies computed for different approaches on the easy and hard multiple-choice answering task, and the filtered hard question set. CCA, nCCA, and CNN+LSTM are trained on the whole image representation for each type of question. nCCA(box) is trained and evaluated on ground-truth bounding-boxes from COCO segmentations. nCCA(all) trains a single embedding using all question types.}\label{table:results1}
\end{table}

\begin{table}[t!]
\centering
\scriptsize
\begin{tabular}{ | l | c | c  c  c | c c c |}
\cline{3-8}
\multicolumn{2}{c}{}&\multicolumn{3}{|c}{\bfseries Easy Task}& \multicolumn{3}{|c|}{\bfseries Hard Task}\\ \hline
& \multirow{2}{*}{\#Q} & \multirow{2}{*}{nCCA} & nCCA & nCCA &  \multirow{2}{*}{nCCA} & nCCA & nCCA \\ &&& (bbox) &(dbox) & & (bbox) & (dbox) \\ \hline
6. obj attr & 2021 & 47.6\% & 53.6\% & 51.4\% &   43.9\%& 47.9\% & 45.2\% \\ \hline
9. per attr & 4206 & 50.2\% & 55.4\% & 51.2\% &  40.0\% & 47.0\% & 43.3\% \\ \hline
\end{tabular}
\caption{Multiple-choice answering using automatic detection for 42 object/person categories.
``bbox'' denotes ground-truth bounding box and ``dbox'' denotes detected bounding box.}
\label{table:detection}
\end{table}

Although this approach is quite simple, it allows us we make two interesting
observations. First, Madlibs outperforms MS COCO on all types of
multiple-choice questions. If Madlibs and MS COCO descriptions provided the
same information, we would expect their performance to be
comparable. Presumably the performance increase for Madlibs is due to the
coverage of targeted descriptions compared to MS COCO's sentences that describe
the overall image content more generally.  Second, the automatically generated
descriptions from the pre-trained CNN+LSTM perform much worse than the actual
MS COCO descriptions, despite doing quite well on general image description
generation (The BLEU-1 score of CNN+LSTM, 0.67, is near human agreement 0.69 on
MS COCO~\cite{vinyals2014show}).

\vspace{-.2cm}
\section{Experiments}
\label{sec:experiments}
\vspace{-.1cm}

In this section we evaluate a series of methods on the Visual Madlibs Dataset
for the targeted natural language generation and multiple-choice question
answering tasks, introduced in Sec.~\ref{sec:tasks}.  As methods, we evaluate
simple joint-embedding methods -- canonical correlation analysis (CCA) and
normalized CCA (nCCA)~\cite{gong2014multi} -- as well as a recent deep-learning
based method for image description generation --
CNN+LSTM~\cite{vinyals2014show}.  We train these models on 80\% of the images
in the MadLibs collection and evaluate their performance on the remaining 20\%.

In our experiments we extract image features using the VGG Convolutional Neural
Network (CNN)~\cite{simonyan2014very}. This model has been trained on the
ILSVRC-2012 dataset to recognize images depicting 1000 object classes, and
generates a 4,096 dimensional image representation. On the sentence side, we
average the Word2Vec of all words in a sentence to obtain a 300 dimensional
representation.  



CCA is an approach for finding a joint embedding between two multi-dimensional
variables, in our case image and text vector representations.  In an attempt to
increase the flexibility of the feature selection and for improving
computational efficiency, Gong \emph{et~al.}~\cite{gong2014multi} proposed a
scalable approximation scheme of explicit kernel mapping followed by dimension
reduction and linear CCA. In the projected latent space, the similarity is
measured by the eigenvalue-weighted normalized correlation. This method, nCCA,
provides high-quality retrieval results, improving over the original CCA
performance significantly~\cite{gong2014multi}.

We train CCA and nCCA models for each question type separately using the
training portion of the Visual Madlibs Dataset.  These models allow us to map
from an image representation, to the joint-embedding space, to vectors in the
Word2Vec space, and vice versa. For targeted generation, we map an image to the
joint-embedding space and then choose the answer from the training set text
that is closest to this embedded point. In order to answer a multiple-choice
question we embed each multiple choice answer, and then select the answer who's
embedding is closest to image.

Following the recent ``Show and Tell'' description generation
technique~\cite{vinyals2014show} (using an implementation
from~\cite{karpathy2014deep}), we train a CNN+LSTM model for each question type
on the Visual Madlibs training set.  This approach has
demonstrated state of the art performance on generating general natural
language descriptions for images.  These models directly learn a mapping from
an image to a sequence of words which we can use to evaluate the targeted
generation task.  Note that we input the words from the prompt,
e.g., The chair is, and then let the CNN+LSTM system generate the remaining
words of the description\footnote{The missing entries for questions 7 and 12
are due to this priming failing for a fraction of the questions.}.  For the
multiple choice task, we compute cosine similarity between Word2Vec
representations of the generated description and each question answer and 
select the most similar answer.

\begin{figure*}[t!]
\centering
\includegraphics[width=0.9\textwidth]{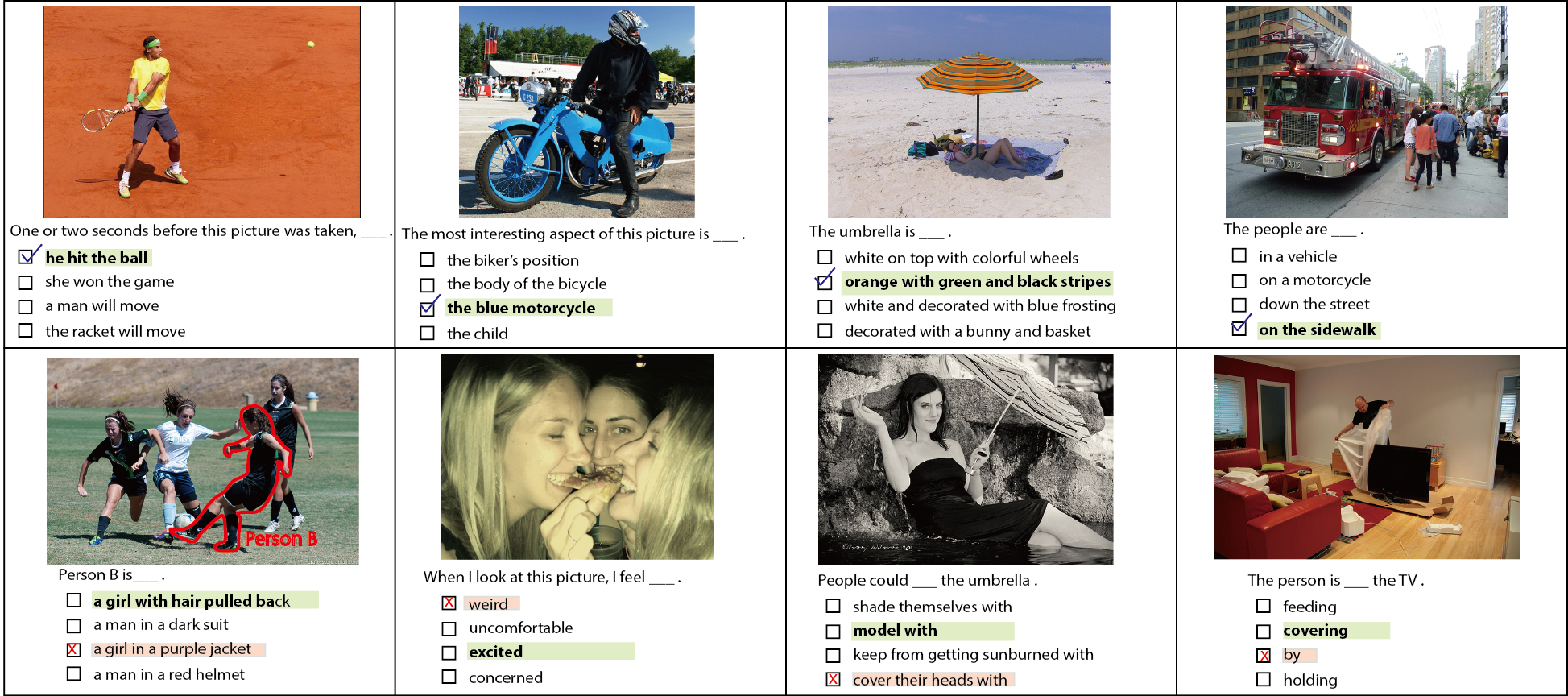}
\caption{Some example question-answering results from nCCA. First row shows correct choices. Second row shows incorrect choices.}\label{fig:results1}
\end{figure*}

\begin{table*}
\centering
\scriptsize
\begin{tabular}{| l | c  c  c | c  c c |}
\hline 
&\multicolumn{3}{c}{\bfseries BLEU-1} & \multicolumn{3}{|c|}{\bfseries BLEU-2}\\
& nCCA & nCCA(bbox )&CNN+LSTM & nCCA & nCCA(box) & CNN+LSTM \\ \hline
1. scene & 0.52 & $-$ & 0.62 & 0.17 & $-$ & 0.19 \\ \hline
2. emotion & 0.17 & $-$ & 0.39 & 0 & $-$ & 0 \\ \hline
3. future & 0.38 & $-$ & 0.32 & 0.12 & $-$ & 0.08 \\ \hline
4. past & 0.39 & $-$ & 0.42 & 0.12 & $-$ & 0.11 \\ \hline
5. interesting & 0.49 & $-$ & 0.51 & 0.14 & $-$ & 0.15 \\ \hline 
6. obj attr & 0.28 & 0.36 & 0.45 & 0.02 & 0.02 & 0.01 \\ \hline
7. obj aff & 0.56 & 0.60 & $-$ & 0.10 & 0.11 & $-$ \\ \hline
8. obj pos & 0.53 & 0.55 & 0.71 & 0.24 & 0.25 & 0.50 \\ \hline
9. per attr & 0.26 & 0.29 & 0.55 & 0.06 & 0.07 & 0.25 \\ \hline
10. per act & 0.47 & 0.41 & 0.52 & 0.14 & 0.11 & 0.22 \\ \hline
11. per loc & 0.52 & 0.46 & 0.64 & 0.22 & 019 & 0.39 \\ \hline
12. pair rel & 0.46 & 0.48 & $-$ & 0.07 & 0.08 & $-$ \\ \hline
\end{tabular}
\caption{BLEU-1 and BLEU-2 computed on Madlibs testing dataset for different approaches.}\label{table:bleuscores}
\end{table*}

\subsection{Discussion of results}
\vspace{-.1cm}
Table~\ref{table:results1} shows accuracies of each algorithm on the easy and
hard versions of the multiple-choice task.  Fig.~\ref{fig:results1}, shows
example correct and wrong answer choices.  There are several interesting
observations we can make.  First, training nCCA on all types of question
together, labeled as nCCA(all), is helpful for the easy variant of the task,
however it is less useful on the ``fine-grained'' hard version of the task.
Second, extracting visual features from the bounding box of the relevant
person/object yields higher accuracy for predicting attributes, but not for
other questions.  Based on this finding, we try answering the attribute
question using automatic detection methods.  The detectors are trained on
ImageNet using R-CNN~\cite{girshick14CVPR}, covering 42 MS COCO categories.  We
observe similar performance between ground-truth and detected bounding boxes
in Table~\ref{table:detection}.

As an additional experiment we ask humans to answer the multiple choice task,
with 5 Turkers answering each question.  We use their results to filter out a
subset of the hard multiple-choice questions where at least 3 Turkers choose
the correct answer.  Results of the methods on this subset are shown in
Table~\ref{table:results1} bottom set of rows.  These results show the same
pattern as on the unfiltered set, with slightly higher accuracy.

Table~\ref{table:bleuscores} shows BLEU-1 and BLEU-2 scores for targeted
generation.  Although the CNN+LSTM models we trained on Madlibs were not quite
as accurate as nCCA for selecting the correct multiple-choice answer, they did
result in better, sometimes much better, accuracy (as measured by BLEU scores)
for targeted generation.

\vspace{-.2cm}
\section{Conclusions}
\label{sec:conclusions}
\vspace{-.2cm}
We have introduced a new fill-in-the blank strategy for targeted natural
language descriptions and used this to collect a Visual Madlibs dataset.
Our analyses show that these descriptions are usually more detailed
than generic whole image descriptions.  We also introduce a targeted natural
language description generation task, and a multiple-choice question answering
task, then train and evaluate joint-embedding and generation models. 
Data produced by this paper will be publicly released upon 
acceptance.

\section*{Acknowledgement}
We thank the vision and language community for feedback regarding this dataset, especially Julia Hockenmaier, Kate Saenko, and Jason Corso. This research is supported by NSF Awards \#1417991, 1405822, 144234, and 1452851, and Microsoft Research.

{\small
\bibliographystyle{ieee}
\bibliography{egbib}
}

\end{document}